\documentclass[10pt,twocolumn,letterpaper]{article}

\usepackage{wacv}
\usepackage{times}
\usepackage{epsfig}
\usepackage{graphicx}
\usepackage{amsmath}
\usepackage{amssymb}
\usepackage{subfigure}
\usepackage{dsfont, booktabs}

\newcommand{\modelnamelong}[0]{\textit{Joint Event Detection and Description Network (JEDDi-Net)~}}
\newcommand{\modelnamelongnospace}{\textit{Joint Event Detection and Description Network (JEDDi-Net)}}
\newcommand{\modelnamelongdot}[0]{\textit{Joint Event Detection and Description Network (JEDDi-Net).}}
\newcommand{\modelname}[0]{JEDDi-Net}
\newcommand*{\affmark}[1][*]{\textsuperscript{#1}}

\newcommand{\specialcell}[2][c]{%
  \begin{tabular}[#1]{@{}l@{}}#2\end{tabular}}

\wacvfinalcopy 

\def\wacvPaperID{581} 

\ifwacvfinal\fi
\begin{document}

\title{Joint Event Detection and Description in Continuous Video Streams}

  
\author{Huijuan Xu\affmark[1]\\
\and
Boyang Li\affmark[2]\\
\and
Vasili Ramanishka\affmark[1]\\
\and
Leonid Sigal\affmark[3]\\
\and
Kate Saenko\affmark[1]\\
\and
{
\affmark[1]Boston University \qquad \affmark[2]Baidu Research  \qquad \affmark[3]University of British Columbia
}\\
\and
{\tt\small \affmark[1]\{hxu, vram, saenko\}@bu.edu, \affmark[2]boyangli@baidu.com, \affmark[3]lsigal@cs.ubc.ca}
}

\maketitle
\ifwacvfinal\thispagestyle{empty}\fi

\begin{abstract}
Dense video captioning is a fine-grained video understanding task that involves two sub-problems: localizing distinct events in a long video stream, and generating captions for the localized events.
We propose the \modelnamelongnospace, which solves the dense video captioning task in an end-to-end fashion. 
  Our model continuously encodes the input video stream with three-dimensional convolutional layers, proposes variable-length temporal events based on pooled features, and generates their captions.
 Proposal features are extracted within each proposal segment through 3D Segment-of-Interest pooling from shared video feature encoding.
 In order to explicitly model temporal relationships between visual events and their captions in a single video, we also propose a two-level hierarchical captioning module that keeps track of context.
  On the large-scale ActivityNet Captions dataset, \modelname~demonstrates improved results as measured by standard 
  metrics. We also present the first dense captioning results on the TACoS-MultiLevel dataset. 
\end{abstract}

\section{Introduction}

\begin{figure}[t]
\centering
\includegraphics[width=0.99\linewidth]{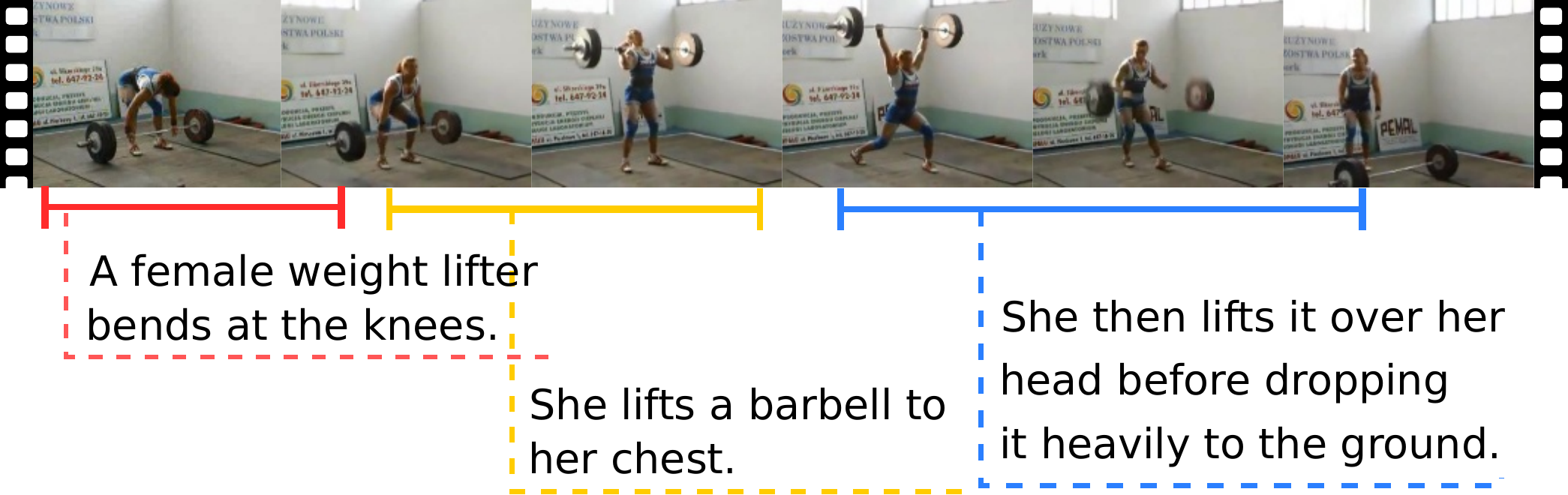}
\vspace{0.1in}
\caption{An example (from ActivityNet Captions~\cite{krishna2017dense}) of the challenges posed by the dense video captioning task. A successful model must detect the time window of each event, which significantly affects the content of predicted captions. The sequential relationship between the three activities in weight lifting suggests that visual and language contexts play a crucial role in this task.
}
\label{fig:vdp-task}
\vskip -0.07in
\end{figure}

The goal of automatic video description is to tell a story about events happening in a video. While early video description methods produced captions for short clips that were manually segmented to contain a single event of interest~\cite{Donahue_2015_CVPR,venugopalan15iccv}, more recently \textit{dense video captioning}~\cite{krishna2017dense} has been proposed to both segment distinct events in time and describe them in a series of coherent sentences. Figure~\ref{fig:vdp-task} shows an example of this task for a weight-lifting video. This problem is a generalization of dense image region captioning~\cite{densecap,yang2016dense} and has many practical applications, such as generating textual summaries for the visually impaired, or detecting and describing important events in surveillance footage.

There are several key challenges in dense video captioning: accurately detecting the start and end of each event, recognizing the type of activity and objects involved, and translating this knowledge into a fluent natural language sentence. The context of the past and future sentences must also be taken into account to construct coherent stories. In~\cite{krishna2017dense}, the authors proposed using two sets of recurrent neural networks (RNNs). The \textit{proposal} RNN encodes convolutional features from input frames and proposes the start and end time of temporal activity segments. The separate two-layer captioning RNN receives the state vector of each activity proposal and decodes it into a sentence.

One issue with the existing approach~\cite{krishna2017dense} is that using the accumulated state vector of the proposal RNN to represent the visual content of a proposed segment may be inaccurate. Each state vector of the proposal RNN is used to predict a set of variable length temporal proposals, while this set of proposals use the same RNN state vector as proposal feature representation. Instead, we want to more precisely capture the activity feature by considering only the frames within that temporal segment.
Another problem is that the temporal segmentation (i.e., proposal generation) stage and the caption generation stage are separately trained. As a result, errors in sentence prediction cannot be propagated back to temporal proposal generation. However, consider  Figure~\ref{fig:vdp-task}: if the temporal proposal for the sentence \textit{``she then lifts it... before dropping it...''} is shortened by a small amount, it would miss the drop part of the activity, resulting in a wrong caption.

In this work, we present a new 
approach to dense video captioning, the \modelnamelongdot~Our model utilizes three-dimensional convolution to extract video appearance and motion features, which are sequentially passed to the temporal event proposal network and the captioning network. Notably, the entire network is end-to-end trainable, with feature computation and temporal segmentation directly influencing captioning loss. 
For proposal generation and refinement, we adapt the proposal network introduced by the Region Convolutional 3D Network (R-C3D) model~\cite{xu2017r} for activity class detection. The proposal network uses 3D convolutional layers to encode the entire input video buffer and proposes variable-length temporal segments as potential activities. Spatio-temporal features are extracted for proposals using 3D Segment-of-Interest (SoI) pooling from the same convolutional feature maps shared by the proposal stage. The resulting proposal features are passed along to the captioning module. We expect to obtain more semantically accurate captioning using this proposal representation, as compared to using the accumulated RNN state representation for a set of proposals~\cite{krishna2017dense}. 

Our \modelname~also uses a hierarchical recurrent caption generator: the low-level \textit{captioner} RNN generates a sentence based on the current proposal's features and on the context that is provided by the high-level \textit{controller} RNN. The captioning model in~\cite{krishna2017dense} also provided context to its sentence generation LSTM module, in the form of visual features from the past and future weighted by their correlation with the current proposal's features. However, the decoded sentences of preceding proposals may also provide useful context information for decoding the current one.
Thus, inspired by~\cite{yu2016video,krause2016hierarchical}, our proposed hierarchical RNN captioning module incorporates both visual and linguistic context. 
The high-level controller RNN accumulates context from visual features and sentences generated so far, and provides it to the low-level sentence captioning module, which generates the new sentence for the target video segment.

\vspace{0.1in}
\noindent \textbf{Contributions:} \modelname~can efficiently detect and describe events in long sequences of frames, including overlapping events of both long and short duration. We summarize the key contributions of our paper as follows:
1) an end-to-end model for the dense video captioning task which jointly detects events and generates their descriptions (code is released for public use\footnote{Code available: https://github.com/VisionLearningGroup/JEDDi-Net});
2) a novel hierarchical language model that incorporates the visual and language context for each new caption and considers the relationships between events in the video;
3) a large-scale evaluation showing improved results on the ActivityNet Captions dataset~\cite{krishna2017dense}, as well as the first dense video captioning results on the TACoS-MultiLevel dataset~\cite{rohrbach2014coherent}.

\section{Related Work}
\vspace{0.1in}
\noindent
\textbf{Activity Detection in Videos:} 
Over the past few years, the video activity  understanding task has quickly evolved from trimmed video classification~\cite{ji20133d,yue2015beyond,simonyan2014two,wang2013action} to activity detection in untrimmed video, as most real-life videos are not nicely segmented and contain multiple activities.
There are two types of activity detection tasks: spatio-temporal and temporal-only.
Spatio-temporal activity detection~\cite{Weinzaepfel_2015_ICCV,Yu_2015_CVPR} localizes activities within spatio-temporal tubes and requires heavier annotation work to collect the training data, while  temporal activity detection~\cite{escorcia2016daps,ma2016learning,montes2016temporal,shou2016temporal,Singh2016a,yeung2016end} only predicts the start and end times of the activities within long untrimmed videos and classifies the overall activity without spatially localizing people and objects in the frame.
Several language tasks related to activity detection have recently emerged in the literature, including the dense video captioning task, which provides detailed captions for temporally localized events~\cite{krishna2017dense}, and the task of language-based event localization in videos~\cite{gao2017tall,hendricks2017localizing}.

Our model includes a temporal activity proposal module which is inspired by the proposal network introduced by the Region Convolutional 3D Network (R-C3D) model~\cite{xu2017r} for activity class detection.
Instead of employing sliding windows~\cite{shou2016temporal,gao2017turn} or RNN feature encodings~\cite{escorcia2016daps,ma2016learning,montes2016temporal,Singh2016a,yeung2016end,buch2017sst} to generate temporal proposals, we encode the input video segment with a fully-convolutional 3D ConvNet and use 3D SoI pooling to allow feature extraction at arbitrary proposal granularities, achieving significantly higher detection accuracy and providing better proposal features for decoding captions. Computation is saved by using 3D SoI pooling to extract proposal features from the shared convolutional feature encoding of the entire input buffer, compared to sliding window approaches which re-extract features for each window from  raw input frames.

\begin{figure*}[t]
\centering
\includegraphics[width=\linewidth]{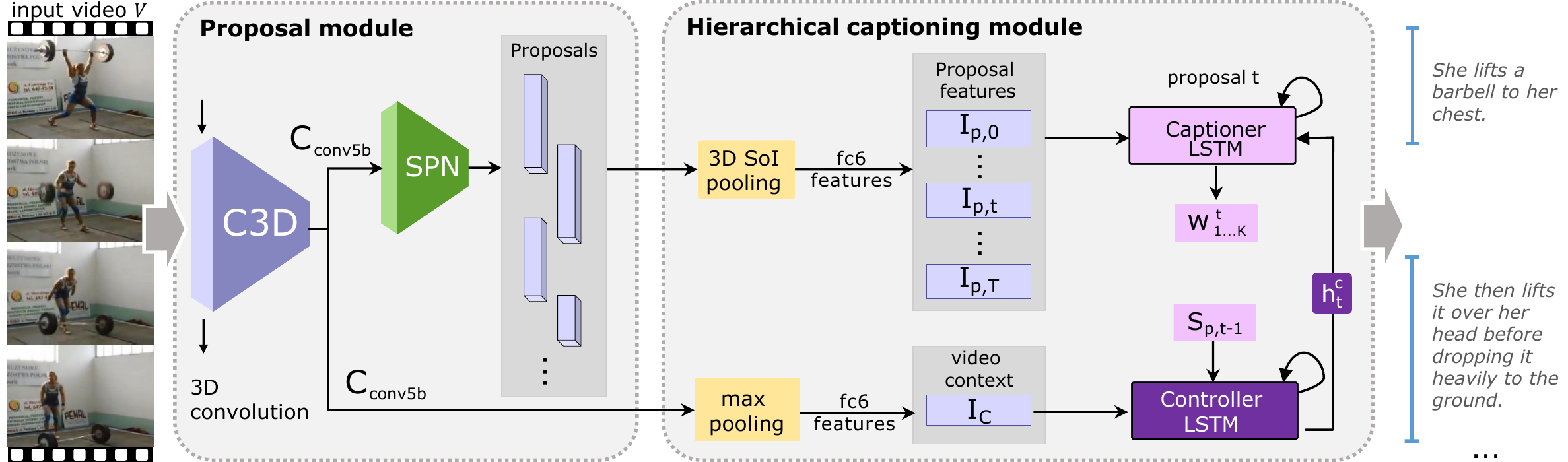} 
\vspace{0.1in}
\caption{The overall architecture of our proposed \modelnamelong consists of two modules. The proposal module (Sec.~\ref{sec:SPN}) extracts features with 3D convolutional layers (C3D) and uses a Segment Proposal Network (SPN) to generate candidate segment proposals (see Fig.~\ref{fig:spn} for details).
The hierarchical captioning module (Sec.~\ref{sec:caption}) contains a controller LSTM to fuse the visual context $I_{c}$ and the decoded language context $S_{p,t-1}$, and provides its hidden state $h^{c}_{t}$ to the captioner LSTM, which decodes the next sentence.
Details of LSTMs are in Fig.~\ref{fig:captioner}.}
\label{fig:architecture}
\end{figure*}

\vspace{0.1in}
\noindent
\textbf{Video Captioning:} 
Early video captioning models (e.g.,~\cite{guadarrama2013youtube2text})  generated a single caption for a trimmed video clip by first predicting the subject, verb and object in the video and then inserting them into a template sentence.
More recent deep models have achieved significantly better trimmed captioning results by using RNNs/LSTMs for language modeling conditioned on CNN features~\cite{venugopalan2014translating,venugopalan15iccv,xu2015multi}.
Attention mechanisms have also been incorporated into RNNs to choose more relevant visual features for decoding captions~\cite{yao2015describing}.

The video paragraph captioning task~\cite{yu2016video} has also been proposed to provide multiple detailed sentence descriptions for long video segments. In contrast to our dense captioning task, video paragraph captioning produces no temporal localization of sentences. 
\cite{yu2016video} proposed a hierarchical RNN to model the language histories when decoding multiple sentences for the video paragraph captioning task, but without explicit visual context modelling. A hierarchical RNN was also applied to image paragraph captioning~\cite{krause2016hierarchical}. However, only the visual context was recorded in the high-level controller layer, and no language history was fed into the controller. 
Hierarchical models have also been applied to natural language processing~\cite{serban2016building}, with \cite{lin2015} proposing a hierarchical RNN language model that integrates sentence history to improve the coherence of documents.
~\cite{krishna2017dense} introduced the dense-captioning task on an ActivityNet-based dataset, and modeled context using attention over past and future visual features.
In this paper, we design a hierarchical captioning module which considers both the visual and language context of the video segment.
Also, in contrast to~\cite{krishna2017dense}, our proposal and captioning modules are jointly trained, with the captioning errors back-propagated to further improve the proposal features and boundaries.

\section{Approach}
\vspace{0.1in}
\noindent \textbf{Overview:} Figure~\ref{fig:architecture} provides an overview of our proposed \modelname~model. 
We assume training data in the form of a video $V$, which contains a number of ground truth segments. For each 
segment, we have its 
 center position $c^{*}$ and length $l^*$ as well as the words in its caption $\{w_{k}\}_{k=1\ldots K}$.
The model consists of two main components: a segment proposal module and a captioning module. 

The \textit{Proposal Module} encodes all input frames in $V$ using a 3D convolutional network (C3D). Based on the features obtained from the layer $conv5b$, the Segment Proposal Network (SPN) proposes temporal segments, classifies them as either potential events for captioning or background, and regresses their temporal boundaries.
The C3D features $C_{conv5b}$ for the video $V$ are also encoded via max-pooling as video context $I_c$, which is utilized in captioning. 

The \textit{Hierarchical Captioning Module} generates a caption for the $t^\text{th}$ proposal, $t=1,\ldots,T$.
This module is composed of a caption-level controller network and a word-level sentence decoder network, both implemented with LSTMs.
The controller network takes the video context vector $I_c$ and the encoding of the previous sentence $S_{p,t-1}$ and provides a single context vector $h^c_{t}$ as a summary of both visual and linguistic context. The word-level decoder network takes as input the current proposal's features $I_{p,t}$ and the context vector $h^c_{t}$ and generates the words $w^t_k$ one by one. 
The entire network is trained end-to-end with three jointly optimized loss functions, including the proposal classification loss, the regression loss on the proposal's center and length, and cross-entropy loss for word prediction. Secs.~\ref{sec:SPN} and \ref{sec:caption} introduce the segment proposal and the hierarchical captioning modules, and Sec.~\ref{sec:optimization} explains the end-to-end optimization strategy.

\vspace{0.1in}
\subsection{Proposal Module}
\label{sec:SPN}
\vspace{0.1in}
\noindent
\textbf{Video Feature Representation}: 
The feature encoding of the input video should extract semantic appearance and dynamic features and preserve temporal sequence information.
We employ the C3D architecture~\cite{tran2015learning} to encode the input frames in a fully-convolutional manner. C3D consists of eight convolutional layers (from \texttt{conv1a} to \texttt{conv5b}). Convolution and pooling in spatiotemporal space allows us to retain temporal sequence information within the input video. 
We represent the sequence of $\mathcal{L}$ RGB video frames of height $\mathcal{H}$ and width $\mathcal{W}$ as  $V \in \mathbb{R}^{3\times \mathcal{L} \times \mathcal{H}\times \mathcal{W}}$.
The C3D convolutions encode $V$ into feature maps $C_{conv5b}\in \mathbb{R}^{512\times \frac{\mathcal{L}}{8} \times \frac{\mathcal{H}}{16}\times \frac{\mathcal{W}}{16}}$ ($512$ is the channel dimension of the layer \texttt{conv5b}). These feature maps are used to produce the proposal features $I_{p,t}$ and video-level visual context $I_c$.

\begin{figure}[t]
\vspace{0.15in}
\centering
\includegraphics[width=0.99\linewidth]{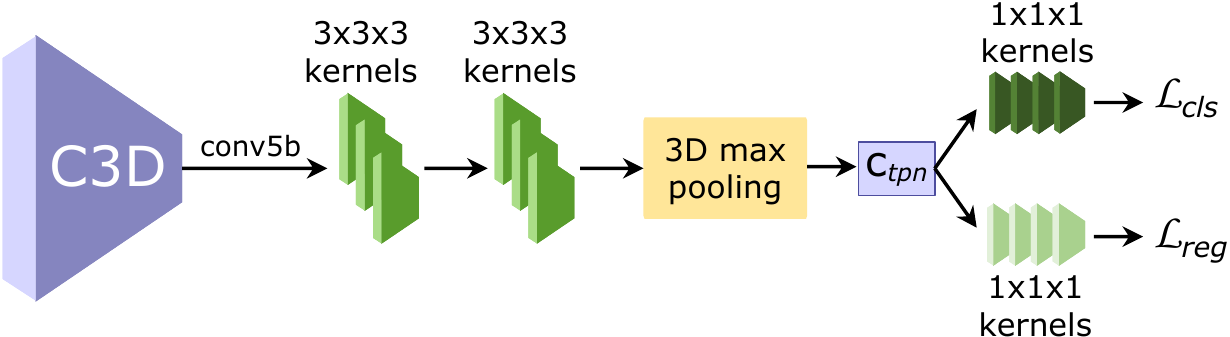}
\vspace{0.2in}
\caption{The structure of the Segment Proposal Network. (Sec.~\ref{sec:SPN})}
\label{fig:spn}
\end{figure}

\vspace{0.1in}
\noindent
\textbf{Segment Proposal Network (SPN)}: 
In this step, we predict the activity proposals' start and end times.  
The accuracy of the proposals' boundary will affect the proposal feature encoding, and will further affect the decoded captions, especially for short activities.
To obtain feature vectors $C_{tpn}\in \mathbb{R}^{512\times \frac{\mathcal{L}}{8} \times 1\times 1}$ for predicting proposals at each of L/8 time points,
we add two 3D convolutional filters with kernel size $3\!\times \!3\!\times \!\!3$ on top of $C_{conv5b}$, followed by a 3D max-pooling filter to remove the spatial dimension.
Proposed segments are predicted around a set of anchor segments~\cite{xu2017r}.
Based on the $512$-dimensional feature vector at each temporal location in $C_{tpn}$, we predict a relative offset $\{\delta c_{i},\delta l_{i}\}$ to the center location and the length of each anchor segment $\{c_{i},l_{i}\}_{i=1\cdots R}$, as well as a binary label indicating whether the predicted proposal contains an activity or not. This is achieved by adding two $1\!\times \!1\!\times \!\!1$ convolutional layers on top of $C_{tpn}$. A detailed diagram of the Segment Proposal Network (SPN) is shown in Figure~\ref{fig:spn}.

\vspace{0.1in}
\noindent
\textbf{Training}:
To train the binary proposal classifier in the segment proposal network, we need a training set with positive and negative examples.
Only positive examples contribute to the proposal regression loss. The ground truth segments' center location and length are transformed with respect to the positive anchor segments using Eq~(\ref{eq:transformation}).
We assign an anchor segment a positive label if it 1) overlaps with some ground-truth activity with temporal Intersection-over-Union (tIoU) higher than 0.7, or 2) has the highest tIoU overlap with some ground-truth activity.
If the anchor has tIoU overlap lower than 0.3 with all ground-truth activities, then it is given a negative label. All others are held out from training.
We sample balanced batches with a positive/negative ratio of $1\!:\!1$.

We train the SPN network by jointly optimizing both the binary proposal classification and proposal boundary regression. For the $i^\text{th}$ anchor segment, $\{ c_i, l_i\}$ denotes the center and the length of the segment and $\hat{a}_i$ denotes the predicted probability.
The corresponding ground truth labels are $a^*_i$, $c^*_i$, and $l^*_i$. 
Ground truth segments are transformed with respect to positive anchor segments following the equations below:
\begin{equation}
\begin{gathered}
  \delta c_i^* = (c_i^* - c_i) / l_i \\
  \delta l_i^* = log(l_i^* / l_i).
\end{gathered}
\label{eq:transformation}
\end{equation}
SPN predicts the offset $\delta \hat{c}_i$ and $\delta \hat{l}_i$. 
The cross-entropy loss, denoted as $\mathcal{L}_{cls}$, is used for binary proposal classification. The smooth L1 loss~\cite{girshick2015fast}, $\mathcal{L}_{reg}$, is used for proposal boundary regression and defined as 
\begin{equation}
\mathcal{L}_{reg}(x) = \mathds{1}(|x|<1) 0.5 x^2 + \mathds{1}(|x|\ge 1) (|x|- 0.5 )
\end{equation}
where $\mathds{1}(\cdot)$ is the indicator function. 
The joint loss function is given by
\small
\begin{equation}
    \hspace{-2mm} \mathcal{L}_{spn} = \frac{1}{M} \sum\limits_{i=1}^{M} \mathcal{L}_{cls} (\hat{a}_i, a_i^*) + a_i^* \left( \mathcal{L}_{reg} (\delta \hat{c}_i \scalebox{1.0}[1.0]{-} \delta c_i^*)
    +  \mathcal{L}_{reg}(\delta \hat{l}_i\scalebox{1.0}[1.0]{-}\delta l_i^*) \right)
\label{eq:loss1}
\end{equation}
\normalsize
where $M$ stands for the number of sampled proposals in the training batch.

At test time, we perform the inverse transformation of Eq~(\ref{eq:transformation}) to find the center and length of predicted proposals. Then, the proposals are refined via Non-Maximum Suppression (NMS) with a tIoU threshold of 0.7.

\vspace{0.1in}
\subsection{The Hierarchical Captioning Module}
\label{sec:caption}

\vspace{0.1in}
\noindent
\textbf{Proposal Feature Encoding:} 
To compute a visual representation of each proposed event for the captioning module, we encode predicted proposals into features ${I_{p,t}}$. In order to encode variable-length proposals, we adopt 3D SoI Pooling, which divides the shared feature map $C_{conv5b}$ equally into bins, performs max-pooling within each bin, and further feeds it through the fc6 layer of the C3D network~\cite{tran2015learning}.
To represent visual context, we encode the entire input video segment $V$ as a vector $I_C$ using a max pooling layer and the shared fc6 layer.

\begin{figure}[t]
\centering
\includegraphics[width=0.99\linewidth]{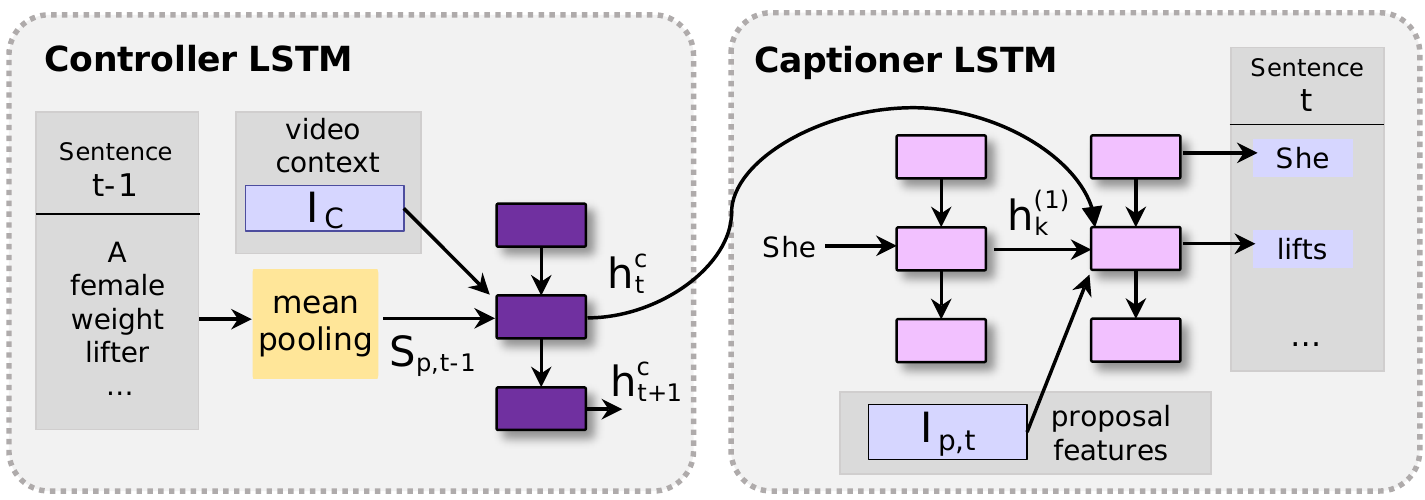}
\vspace{0.1in}
\caption{The structure of LSTMs in the hierarchical captioning module. (Sec.~\ref{sec:caption})}
\label{fig:captioner}
\end{figure}

\vspace{0.1in}
\noindent
\textbf{Controller LSTM:} 
To model context
between the generated caption sentences, we adopt a hierarchical LSTM structure. 
The high-level Controller LSTM encodes the visual context and sentence decoding history. 
The low-level Captioning LSTM decodes every proposal into a caption word by word, while being aware of visual and language context. Figure \ref{fig:captioner} illustrates this hierarchical structure. 

The controller is a single layer LSTM which accepts the visual context vector $I_c$ and the caption sentence of the previous proposal, encoded as $S_{p,t-1}$. 
The LSTM hidden state $h^c_t$ of the controller encodes the visual context and the language history, and serves as a topic vector, which is fed to the sentence captioning LSTM.
The recurrence equations for the controller are given as:
\begin{eqnarray}
\begin{bmatrix}
f^c_t \\
i^c_t \\
o^c_t \\
\end{bmatrix} & = & \sigma \left(
\begin{bmatrix}
W^c_f \\
W^c_i \\
W^c_o \\
\end{bmatrix} 
\begin{bmatrix}
S_{p,t-1} \\
I_c \\
h^c_{t-1} \\
\end{bmatrix} + \begin{bmatrix}
b^c_f \\
b^c_i \\
b^c_o \\
\end{bmatrix}
\right) \\
\widetilde{c_t}^c & = & \text{tanh}\left(W^c_{c} \begin{bmatrix}
S_{p,t-1} \; I_c \; h^c_{t-1}
\end{bmatrix}  + b^c_c \right) \\
c^c_t &=& i^c_t \otimes \widetilde{c_t}^c + f^c_t \otimes c^c_{t-1} \\
h^c_t & = & o^c_t \otimes \text{tanh}(c^c_t)
\end{eqnarray}
where $\otimes$ is  component-wise multiplication.

The first hidden state $h^c_0$ and the first sentence feature $S_{p,0}$ are initialized to zero. Thus, only visual features are used for decoding the first proposal. 
At training time, ground truth segments are sorted by ascending end time and their captions' encodings are fed to the controller LSTM in sequence. At test time, we sort the predicted proposals by their end times and decode them sequentially. 
For the encoding of the previous caption $S_{p,t-1}$, we experimented with two encoding methods:
the mean-pooling of word vectors, or the last hidden state of the captioner LSTM. Preliminary experiments found no obvious differences in performance, so we adopt mean-pooling for simplicity.

\vspace{0.1in}
\noindent
\textbf{Sentence Captioning LSTM:}  
We design a two-layer LSTM network for decoding proposals into captions. The first layer focuses on learning the word sequence encoding and the second layer focuses on learning the fusion of visual and language information and context.
Each sentence is given a maximum length $K$, and is padded if it is shorter than $K$ words. 
As input to the first layer, each word is represented using word vectors $w^{t}_{k}$. The hidden state of the first layer LSTM, $h^{(1)}_{k}$, is fed to the second layer LSTM, along with the proposal features $I_{p,t}$ and the context vector $h^c_{t}$ from the controller LSTM. 
The recurrence equations for the second layer LSTM are given as follows:
\begin{eqnarray}
\begin{bmatrix}
f_k \\
i_k \\
o_k \\
\end{bmatrix} &=& \sigma \left( \begin{bmatrix}
W_{f} \\
W_{i} \\
W_o \\ 
\end{bmatrix} 
\begin{bmatrix}
h^{(1)}_{k}  \\
I_{p,t} \\
h^c_t  \\
h^{(2)}_{k-1} 
\end{bmatrix} + 
\begin{bmatrix}
b_{f}  \\
b_{i} \\
b_{o}  \\
\end{bmatrix} \right) \\
\widetilde{c_k} & = & \text{tanh}\left(W_{c} \begin{bmatrix}
h^{(1)}_{k} \; I_{p,t} \; h^c_t \; h^{(2)}_{k-1} 
\end{bmatrix} + b_{c}\right) \\
c_k & = & i_k \otimes \widetilde{c_k} + f_k \otimes c_{k-1} \\
h_k & = & o_k \otimes \text{tanh}(c_k)
\end{eqnarray}

The hidden state $h^{(2)}_k$ goes through a softmax and is used to predict the word at the $k^\text{th}$ position in the caption. We optimize the normalized log likelihood over all $T$ ground truth proposals and all $K$ unrolled timesteps in the sentence captioning module: 
\small
\begin{equation}
    \hspace{-2mm} \mathcal{L}_{caption} =  -\frac{1}{KT} \sum\limits_{t,k} \log P(w^{t}_{k}| I_{p,t}, h^c_t ,w^{t}_{1},...,w^{t}_{k-1}).
\label{eq:loss2}
\end{equation}
\normalsize

\vspace{0.1in}
\subsection{End-to-End Optimization}
\label{sec:optimization}
\modelname~can be trained end-to-end with the proposal and hierarchical captioning modules optimized jointly. The overall loss is as follows; we set $\lambda=1$.
\small
\begin{equation}
    \hspace{-2mm} \mathcal{L}_{total} = \mathcal{L}_{spn} + \lambda \mathcal{L}_{caption} 
\label{eq:loss3}
\end{equation}
\normalsize

Our end-to-end training allows us to propagate gradient information back to the underlying C3D network and optimize the convolutional filters for better proposal features and visual context encoding. 
In activity detection, multiple positive and negative proposals are generated according to tIoU thresholds with ground truth segments in a single video and selectively form balanced training mini-batches.
In dense captioning, however, a video contains only a few ground-truth captions. 
Further, the same captions always appear together in the same mini-batch with one video as input during end-to-end training. We find the lack of diversity to severely disrupt proper optimization.

We propose a more effective training strategy.
We first extract intermediate ground truth segment features from the pretrained SPN and C3D classification networks. We then shuffle these and form a relatively large training batch with diverse captions to pre-train the captioning module.
After pretraining, the entire network is trained end-to-end following the conventional strategy with a reduced learning rate.
In the experimental section, we show substantial performance improvements after end-to-end training compared to the separately trained models.
In the next section, we present experimental results illustrating the benefits of end-to-end training on both proposal prediction and caption generation.

\section{Experiments}

We evaluate \modelname~on the large-scale ActivityNet Captions dataset proposed by~\cite{krishna2017dense}.
For proposal evaluation, we use the conventional Area Under the AR vs AN curve (AUC) with tIoU threshold 0.8.
When evaluating captions, we follow \cite{krishna2017dense} by computing the average precision (BLEU, METEOR, CIDEr and ROUGE\_L) across tIoU thresholds of 0.3, 0.5, 0.7, 0.9 for the top 1000 proposals. 
In addition, we report results on the TACoS-MultiLevel dataset~\cite{rohrbach2014coherent}. 

\vspace{0.1in}
\subsection{Experiments on the ActivityNet Captions}
\label{exp:ActivityNet}

\vspace{0.1in}
\noindent
\textbf{Dataset and Setup:}
The ActivityNet Captions dataset~\cite{krishna2017dense} with around 20k videos are split into training, validation and testing with a 50\%/25\%/25\% ratio.
Each video contains at least two ground truth segments and each segment is paired with one ground truth caption.
We keep all the words that appear at least 5 times.
The height $\mathcal{H}$ and width $\mathcal{W}$ of all input frames are set to 112 each following~\cite{tran2015learning}. We set the number of frames $\mathcal{L}$ to 768, breaking the arbitrary length input video into 768 frame chunks and zero-padding if necessary.
The maximum caption length is set to 30, which covers over 97\% of captions in the training set.
We sample frames at 3 fps and set the number of anchor segment scales to be 36 to generate candidate proposals\footnote{Specifically, we chose the following anchor scales based on cross-validation - [1,2,3,4,5,6,7,8,10,12,14,16,20,24,28,32,40,48,56,64,66,68,\\70,72,74,76,78,80,82,84,86,88,90,92,94,96].}. 
In the hierarchical captioning module, we set the hidden state dimension to 20 in the controller LSTM and 512 in the captioner LSTMs.

We train the SPN using the temporal annotation of ground truth segments in the ActivityNet Captions dataset with Sports-1M pretrained C3D weight initialized~\cite{tran2015learning}.
We also extract fc6 features for ground truth proposals from pretrained SPN, shuffle the proposal features and paired ground truth captions, and form batches of size 32 to train the captioner LSTM from scratch.
The pretrained SPN and captioner LSTM will serve as initialization weights for our end-to-end model. 
We refer our full \modelname~which is jointly trained for SPN and hierarchical captioning modules as `\modelname (joint training with context)'.
After removing the controller LSTM of the hierarchical captioning module in `\modelname (joint training with context)', we refer this ablation model as `\modelname (joint training)'.
To show the effectiveness of end-to-end training in our model, we extract proposal features from the separately trained SPN and decode captions using the separately trained captioner LSTM, and refer to this model as `\modelname (separate training)'.

\begin{table}[!t]
\centering
\caption{Proposal evaluation results on ActivityNet Captions dataset (in percentage). AUC at IoU threshold 0.8 and average AUC at tIoU thresholds $\alpha \in (0.5,0.95)$ with step 0.05 are reported.}
\vspace{0.1in}
\resizebox{\columnwidth}{!}{
\begin{tabular}{l  c   c} 
\toprule
~ & \, $\alpha =0.8$ \, & \, $\alpha \in (0.5,0.95)$ \, \\ \hline
DAP~\cite{krishna2017dense} & 30 & -  \\ 
multi-scale DAP~\cite{krishna2017dense} & 38 & - \\ \hline
pretrain SPN & 57.75 & 57.12 \\ 
\modelname (joint training) & 59.13 & 58.70 \\
\modelname (joint training w/ context) & 58.21 & 58.24 \\ \hline 
\end{tabular} }
\label{res:proposal}
\end{table}

\vspace{0.1in}
\noindent
\textbf{Proposal Evaluation:}
The proposal evaluation result is shown in Table~\ref{res:proposal}.
The dense video captioning model in~\cite{krishna2017dense} uses DAP~\cite{escorcia2016daps} as its proposal network, extends DAP to a multi-scale version and shows improved proposal results in AUC at tIoU 0.8.
Our pretrained SPN model achieves 57.75\% at tIoU 0.8 in AUC, 19.75\% higher than \cite{krishna2017dense}, indicating our superior ability to segment events of interest. 
Following the traditional evaluation of the temporal localization task in ActivityNet, we also report the average AUC result across ten different tIoU thresholds uniformly distributed between 0.5 and 0.95 with 1000 proposals per video. The average AUC for our pretrained SPN is 57.12\%, which is on par with tIoU at 0.8, indicating robust performance of SPN across different tIoUs.

\begin{table*}[!t]
\centering
\caption{Dense video captioning results on ActivityNet Captions dataset (in percentage). The average Bleu\_1-4 (B1-B4), METEOR (M), CIDEr (C) and ROUGE\_L (R) across tIoU thresholds of 0.3, 0.5, 0.7, 0.9 are reported.}
\vspace{0.1in}
\begin{tabular}{l c c c c c c c c} 
\hline
 Model &  B1 & B2  & B3  & B4 & M & C & R \\ \hline
  R. Krishna et al.~\cite{krishna2017dense} (no context) & 12.23 &  3.48 &  2.1 & 0.88 & 3.76 & 12.34 &  - \\ 
 R. Krishna et al.~\cite{krishna2017dense} (with context)  & 17.95 &  7.69 &  3.86 & \bf{2.20} & 4.82 & 17.29 &  - \\ \hline
 \modelname (separate training) & 16.72 &  6.65 &  2.65 & 1.07 & 7.37 & 14.65 &  16.47 \\ 
 \modelname (joint training) & 19.27 &  8.69 &  3.78 & 1.54 & 8.30 & 19.81 &  18.86 \\ 
 \modelname (joint training w/ context) & \bf{19.97} &  \bf{9.10} &  \bf{4.06} & 1.63 & \bf{8.58} & \bf{19.88} &  \bf{19.63} \\ \hline
 \specialcell{\modelname (joint training w/ context)~\\on test server} &  - &  - &  - & - & 8.81 & - &  - \\ \hline
\end{tabular}
\label{tab:res_act_avg}
\vspace{0.1in}
\end{table*}

\begin{table*}[!t]
\centering
\caption{Dense video captioning results at different tIoU thresholds $\alpha$ on ActivityNet Captions dataset (in percentage). The Bleu\_1-4 (B1-B4), METEOR (M), CIDEr (C), and ROUGE\_L (R) at different tIoU thresholds $\alpha$ are reported for our \modelname (joint training with context)~with greedy search decoding.}
\vspace{0.1in}
\begin{tabular}{l c c c c c c c c} 
\hline
$\alpha$ & B1 &  B2  & B3  & B4 & M & C & R \\ \hline
  0.3 & \, 19.72 \, & \, 8.84 \, & \, 4.04 \, & \, 1.65 \, & \, 8.44 \, & \, 13.40 \, & \, 19.80 \, \\ 
  0.5 & 20.31 &  9.26 &  4.22 & 1.71 & 8.75 & 16.53 & 20.41 \\ 
  0.7 & 20.86 &  9.70 &  4.37 & 1.76 & 8.97 & 21.52 & 20.74 \\ 
  0.9 & 19.00 &  8.60 &  3.61 & 1.39 & 8.17 & 28.09 & 17.57 \\ \hline
 avg $\alpha \in (0.3,0.5,0.7,0.9)$ \; & 19.97 &  9.10 &  4.06 & 1.63 & 8.58 & 19.88 &  19.63 \\ \hline
\end{tabular}
\label{tab:res_act_tiou}
\vspace{0.1in}
\end{table*}

\vspace{0.1in}
\noindent
\textbf{Dense Captioning Evaluation:} The average dense video captioning results across four tIoUs using the evaluation code released by~\cite{krishna2017dense} are shown in Table~\ref{tab:res_act_avg}. 
We list the two baseline results from~\cite{krishna2017dense}, the model without visual context and the one with visual context.
Our first `\modelname (separate training)' model without end-to-end training already achieves reasonable results with a METEOR score 2.55\% higher than the best context model in~\cite{krishna2017dense}.
This indicates that our decoded captions are more semantically meaningful and closer to human descriptions. These results further motivate our proposal feature encoding method, which employs 3D SoI pooling directly on the conv features of the input video segment, rather than using the LSTM hidden state for a set of proposals.
Our `\modelname (separate training)' and `\modelname (joint training)' models without context do better than \cite{krishna2017dense}'s `no context' model on all evaluation metrics.
After end-to-end training, both `\modelname (joint training)' and `\modelname (joint training with context)' improve on all evaluation metrics compared to `\modelname (separate training)'. This shows the benefits of joint parameter training for dense video captioning. 
Our `\modelname (joint training with context)' model that incorporates visual and language context further improves all the language evaluation metrics compared to the no context version.

Our full model outperforms the context model in~\cite{krishna2017dense} on all evaluation metrics except for Bleu\_4. In particular, we achieve a 78\% relative improvement on METEOR, the only metric used by the test server. 
The reason for lower BLEU\_4 might be that we did not leverage the power of beam search due to limited computational resources.
We decoded the captions with greedy search (in Table~\ref{tab:res_act_avg}), selecting the most probable word at each timestep.
Experiments in several papers~\cite{Vinyals_2015_CVPR,fang2015captions} show that beam search can improve some evaluation metrics, especially Bleu\_3 and Bleu\_4.

Applying the same \modelname (joint training with context)~on the test server yields an average METEOR score of 8.81\%, 
which is on the same level as the average METEOR score 8.58\% on the validation set. This demonstrates that our model generalizes well to unseen data.

Table~\ref{tab:res_act_tiou} shows all the evaluation metrics for all the four tIoUs in details for our `\modelname (joint training with context)'.
As tIoU $\alpha$ increases from 0.3 to 0.7, Bleu\_1-4, METEOR and ROUGE\_L increase steadily, with the highest scores at $\alpha=0.7$.
The reason might be that our SPN network is trained with tIoU greater than 0.7 as positive examples, and tested with post processed NMS at 0.7.
However, Bleu\_1-4, METEOR and ROUGE\_L decrease significantly at tIoU 0.9, possibly because much fewer proposals have been left for evaluation using the tIoU 0.9 criterion.
The CIDEr metric is consistently improved across tIoU $\alpha$ values from 0.3 to 0.9, which indicates the sensitivity of the CIDEr score to the number of evaluation proposals.
CIDEr measures the diversity of the captions. When a small subset of proposals is kept with higher tIoU, the captions are more diverse and the CIDEr score is higher, and vice versa.

We show two videos with predicted dense captions from \modelname~as qualitative examples from the ActivityNet Captions dataset in Figure~\ref{fig:vis_act}. Our model generates continuous and fluent descriptions of the activities of jumping over the mat and making a cocktail, taking  context into account.
We note that the ground truth caption for segment A in the first video is ``A man is seen...", while our prediction is ``A person is seen...". Though these two  4-grams have the same meaning in this video, such predictions will not be counted as positive in the Bleu\_4 score, indicating a potential reason for the lower value.

\subsection{Experiments on the TACoS-MultiLevel Dataset}
\label{exp:TACoS}

\vspace{0.1in}
\noindent
\textbf{Dataset and Setup:}
The TACoS-MultiLevel dataset~\cite{rohrbach2014coherent} contains cooking videos with the start and end time for captions and activity labels, which can be used for dense video captioning. Compared to ActivityNet Captions, TACoS has more ground truth annotations per video with an average of 284 sentences per video. 
We use the same 143/42 video split for training and testing as in~\cite{yu2016video}.
All words are kept in the vocabulary and the maximum caption length is set to 15. Frames are sampled at 5 fps.
Other settings are identical with the ActivityNet Captions experiments.
We evaluate three ablated models on proposal detection and caption generation.

\begin{table}[!t]
\centering
\caption{Proposal evaluation results on TACoS-MultiLevel dataset, showing AUC at tIoU threshold 0.8 and average AUC at tIoU $\alpha \in (0.5,0.95)$ with step 0.05.}
\vspace{0.1in}
\resizebox{\columnwidth}{!}{
\begin{tabular}{l c  c} 
\hline
~ & \; $\alpha =0.8$ \; & \; $\alpha \in (0.5,0.95)$  \\ \hline
pretrain SPN & 36.88 & 41.90 \\ 
\modelname (joint training) \; & 36.85 & 43.30 \\ 
\modelname (joint training w/ context) \; & 36.31 & 43.23 \\ \hline
\end{tabular} }
\label{res:proposal_tacos}
\end{table}

\begin{table*}[!t]
\centering
\caption{Dense video captioning results on TACoS-MultiLevel dataset (in percentage). The average Bleu\_1-4 (B1-B4), METEOR (M), CIDEr (C) and ROUGE\_L (R) across tIoU thresholds of 0.3, 0.5, 0.7, 0.9 are reported.}
\vspace{0.1in}
\begin{tabular}{l c c c c c c c c} 
\hline
 models &  B1 &  B2  &  B3  &  B4 & M & C & R \\ \hline
  \modelname (separate training) \; &  45.2 &   32.3 &   19.7 &  13.1 &  20.7 &  65.4 &  46.2 \\ 
   \modelname (joint training)  \; & 48.7 &  36.4 &  24.6 & 17.4 & 23.3 & 99.7 & 50.0 \\ 
 \modelname (joint training w/ context)  \; & \textbf{49.2} &  \textbf{37.1} &  \textbf{25.2} & \textbf{18.1} & \textbf{23.9} & \textbf{104.0} & \textbf{50.9} \\ \hline
\end{tabular}
\label{tab:res_tacos_avg}
\vspace{0.1in}
\end{table*}

\vspace{0.1in}
\noindent
\textbf{Results:}
Table~\ref{res:proposal_tacos} shows results of proposal evaluation.
We report the average AUC result across ten tIoU thresholds uniformly distributed between 0.5 and 0.95 for top 1000 proposals per video. 
We also measure the improvement of proposal detection after end-to-end training.
The average AUC for both `\modelname (joint training)' and `\modelname (joint training with context)' improve compared with the pretrained SPN, while AUC at tIoU 0.8 stays almost the same.

Table~\ref{tab:res_tacos_avg} shows results for caption generation averaged across four tIoUs. 
No dense captioning results have been previously reported on this dataset, so ours is the first set of such results. 
The previously reported trimmed video captioning results can be considered as the upper bound for our task on the same annotations, as it is noted in~\cite{krishna2017dense}.
TACoS-MultiLevel dataset~\cite{rohrbach2014coherent} reports a Bleu\_4 value of 27.5\% for trimmed video captioning, which can be seen as the upper bound of our reported Bleu\_4 value with the consideration of tIoU overlaps.
Compared to `\modelname (separate training)', all evaluation metrics for both `\modelname (joint training)' and `\modelname (joint training with context)' improve after end-to-end training, which indicates the benefits of our approach.
Also `\modelname (joint training with context)' further improves all evaluation metrics through modelling of visual and language context in the hierarchical captioning module, compared to `\modelname (joint training)' without explicitly modeling context. 

\begin{figure}[!t]
\centering
\subfigure[ActivityNet Captions dataset]{
\includegraphics[width=0.99\linewidth]{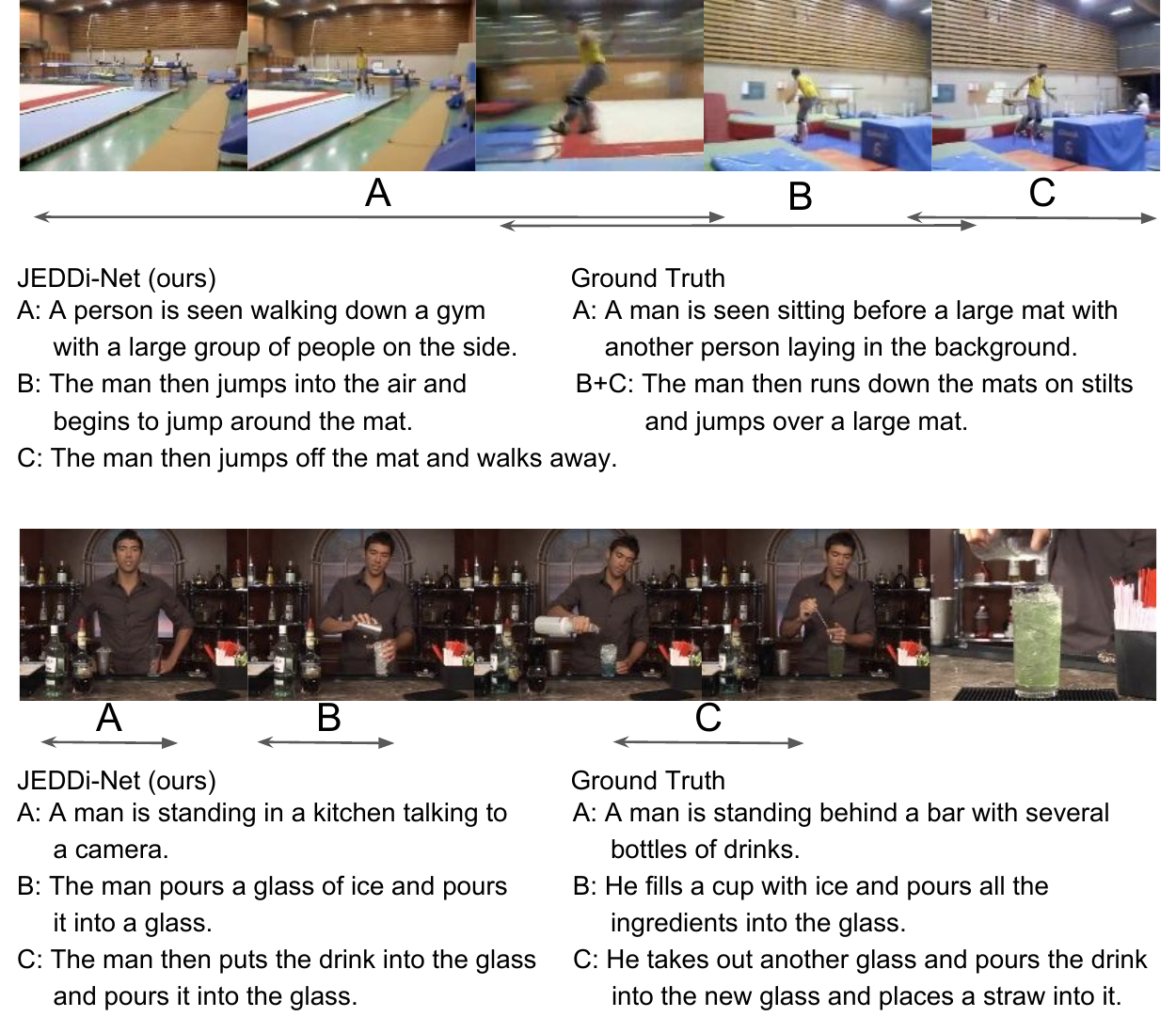}
\label{fig:vis_act}
}
\vspace{0.1in}
\subfigure[TACoS-MultiLevel dataset]{
\label{fig:vis_tacos}
\includegraphics[width=0.99\linewidth]{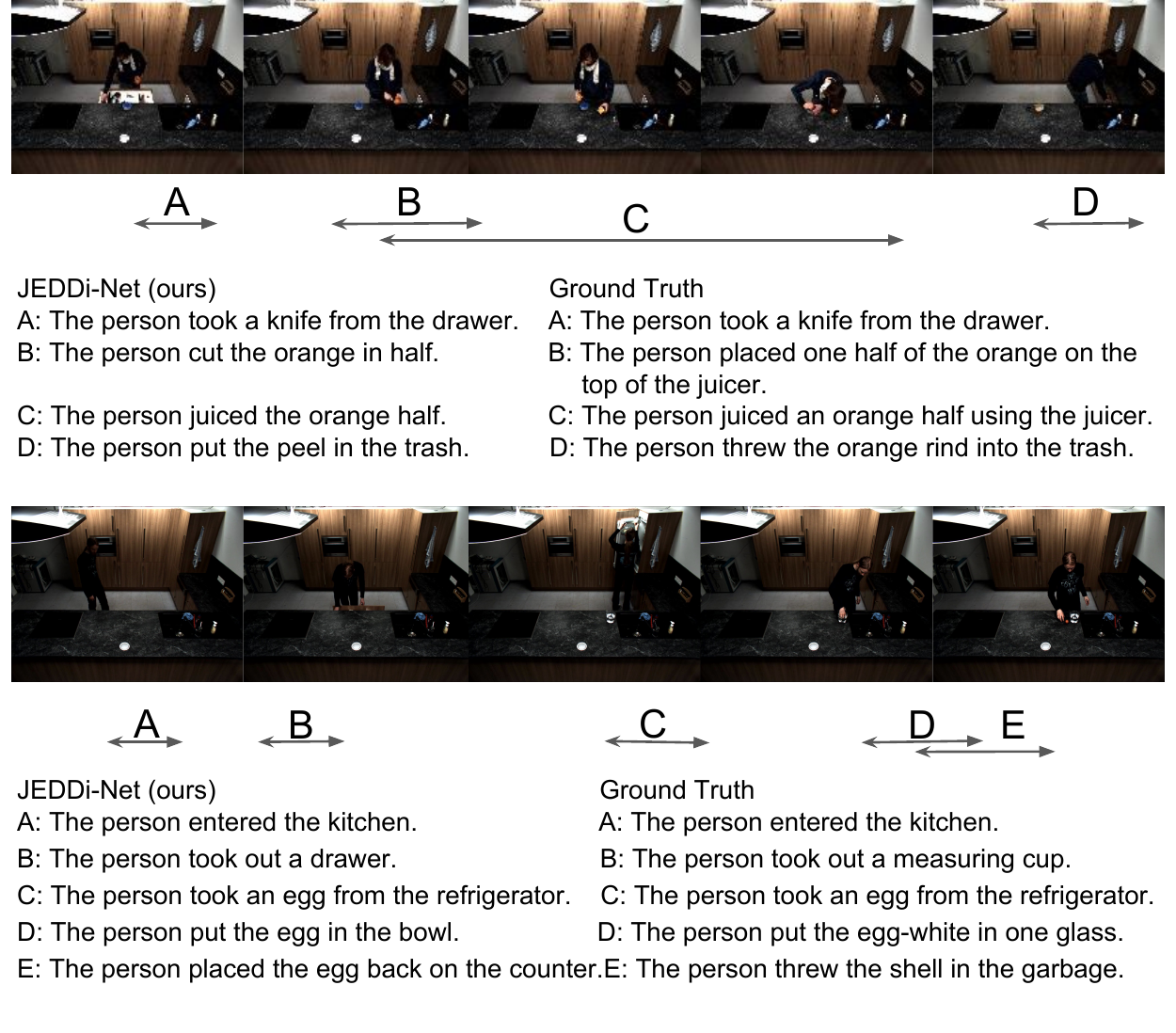}}
\caption
{Qualitative visualization of the predicted dense captions by \modelname~(best viewed in color). Figure \subref{fig:vis_act} and \subref{fig:vis_tacos} show results for two videos each in ActivityNet captions dataset and TACoS-MultiLevel dataset. 
}
\label{fig:qualitative}
\end{figure}

Figure~\ref{fig:vis_tacos} provides two examples of video predictions from TACoS-MultiLevel dataset. 
Though~\modelname~missed some objects in the generated captions like ``a measuring cup'', \modelname~could still provide fine-grained descriptions of certain activities involving small objects such as the orange and the egg. The network likely benefited from learning object representations from the captions in end-to-end training.

\section{Conclusion}
In this paper, we proposed \modelname, an end-to-end deep neural network designed to perform the dense video captioning task, and introduced an optimization strategy for training it end-to-end. The visual and language context is incorporated by the controller in the hierarchical captioning module, to provide context for decoding each proposal rather than training and decoding each proposal independently. 
Our end-to-end framework can be further extended to solve other vision and language tasks, such as natural language localization in videos.

\noindent{\bf Acknowledgements:} Supported in part by IARPA (contract number D17PC00344) and DARPA's XAI program.
\footnote{The U.S. Government is authorized to reproduce and distribute reprints for Governmental purposes notwithstanding any copyright annotation thereon. Disclaimer: The views and conclusions contained herein are those of the authors and should not be interpreted as necessarily representing the official policies or endorsements, either expressed or implied, of IARPA, DOI/IBC, or the U.S. Government.}

{\small
\bibliographystyle{ieee}
\bibliography{egbib}
}

\end{document}